\documentclass[sigconf,authorversion,nonacm]{acmart}

\usepackage{amsmath}
\usepackage{times}
\usepackage{latexsym}
\usepackage{enumitem}
\usepackage[T1]{fontenc}
\usepackage[utf8]{inputenc}

\usepackage{xspace}
\usepackage{todonotes}
\usepackage{booktabs}
\usepackage{soul}
\usepackage{algorithm}
\usepackage{algorithmic}
\usepackage{newfloat}
\usepackage{listings}

\usepackage{booktabs}
\usepackage{graphicx}
\usepackage{subcaption} 
\usepackage{fontawesome} 
\usepackage{comment}
\usepackage{xcolor}
\usepackage{arabtex}
\usepackage{multirow}
\usepackage{makecell}
\usepackage{utf8}
\usepackage[utf8]{inputenc}
\setcode{utf8}
\usepackage[export]{adjustbox}
\usepackage{color, colortbl}
\definecolor{LightCyan}{rgb}{0.95,1,1}

\AtBeginDocument{%
  \providecommand\BibTeX{{%
    \normalfont B\kern-0.5em{\scshape i\kern-0.25em b}\kern-0.8em\TeX}}}

\setcopyright{acmcopyright}
\copyrightyear{2018}
\acmYear{2018}
\acmDOI{XXXXXXX.XXXXXXX}

\acmConference[Conference acronym 'XX]{Make sure to enter the correct
  conference title from your rights confirmation emai}{June 03--05,
  2018}{Woodstock, NY}
\acmPrice{15.00}
\acmISBN{978-1-4503-XXXX-X/18/06}

\begin{document}

\title{ArabGend: Gender Analysis and Inference on Arabic Twitter}



\author{Hamdy Mubarak, Shammur Absar Chowdhury, Firoj Alam}
\affiliation{
  \institution{Qatar Computing Research Institute, HBKU}
  \city{Doha}
  \country{Qatar}
}
\email{{hmubarak, shchowdhury, fialam}@hbku.edu.qa}

\renewcommand{\shortauthors}{Hamdy Mubarak, et al.}

\begin{abstract}
Gender analysis of Twitter can reveal important socio-cultural differences between male and female users. There has been a significant effort to analyze and automatically infer gender in the past for most widely spoken languages' content, however, to our knowledge very limited work has been done for Arabic. In this paper, we perform an extensive analysis of differences between male and female users on the Arabic Twitter-sphere. We study differences in user engagement, topics of interest, and the gender gap in professions. Along with gender analysis, we also propose a method to infer gender by utilizing usernames, profile pictures, tweets, and networks of friends. 
In order to do so, we manually annotated gender and locations for $\sim$166K Twitter accounts associated with $\sim$92K user location, which we plan to make publicly available.\footnote{Please contact authors for data access.} 
Our proposed gender inference method achieve an F1 score of 82.1\%, which is 47.3\% higher than majority baseline. In addition, we also developed a demo and made it publicly available. 
\end{abstract}



\maketitle


\section{Introduction}
\label{sec:introduction}
Demographic information (e.g., age, gender) has proven to be useful in many different decision-making processes such as from business decisions (e.g., personalized online advertising), forensic investigation to policy-making purposes~\cite{li-etal-2016-semi, volkova-etal-2013-exploring, mukherjee-liu-2010-improving,soler-wanner-2016-semi}. For example, social media platforms and e-commerce sites are using customers' gender and other demographic attributes for targeted advertising \cite{9023806}. In the past decade, there have been extensive research efforts to automatically infer demographic attributes of the social media users using their social media footprints (e.g., users' posts, names, and other attributes) \cite{chen2015comparative,volkova2015inferring}. Major research efforts for such attributes inference are mostly done for English, and very little efforts for non-English languages \cite{ciot2013gender}. The research for Arabic demographic inference such as gender is relatively rare for social media users, specifically for Twitter. 
 
With approximately 164 million monthly active users, Twitter is one of the most popular social media platforms in the Arab region~\cite{abdelali2020arabic}. The large volume of tweets produced represents the social and cultural characteristics of the region. Even though there is a large number of Twitter users, however, usage of Twitter differs in volume, topics, and engagement depending on the users' gender role. Another important factor is that social media users often provide misleading demographic information (e.g., name, age, location and marital status), which is highlighted in a survey conducted in the Arab region \cite{salem2017social}. Hence, self-declared information might not be always  reliable. Though some studies argue that the proportion of such misleading self-reported information is relatively lower~\cite{herring2014gender}. While the availability of Twitter data and its large user base provides opportunities to understand such information, however, unfortunately, Twitter does not provide users' gender information \cite{mueller2016gender}. Such factors stress the need to have automatic methods for gender inference, and here our focus is Twitter-sphere for the Arabic region. In addition, there is a gap in the literature in a thorough analysis of Arabic Twitter (e.g., linguistic content) for gender, even though Arabic is a morphologically rich language where linguistic markers are present to distinguish gender roles in many cases (see Section \ref{ssec:arabic_back}).

To address the gap of gender analysis and automatic inference, in this paper, we perform an extensive analysis of Arabic Twitter data where we identify key distinguishing properties of male/female authorship. We experiment with different features to identify the gender of Twitter users. We examine the usage of friendship networks, profile pictures, and textual information such as username, user description, and tweets to classify gender. The contributions of our work are as follows:

\begin{itemize}
	\item We developed a new dataset of $\sim$166K Twitter accounts that are manually annotated for their gender and location, which we plan to make publicly available.
	\item We perform extensive analysis and study how language usage differs based on gender.
	\item We study automatic gender identification of tweets, user accounts, and user descriptions. We also study how profile pictures and networks of friends can influence gender inference models.
	\item Using our models we developed a demo, which we make publicly available. 
\end{itemize}

The rest of the paper is organized as follows: In Section~\ref{sec:related_work}, we provide a brief overview of previous work. We discuss the detail of the dataset in Section~\ref{sec:dataset} and a detail of the annotation in Section~\ref{sec:annotation}. In Section~\ref{sec:analysis}, we present an in-depth analysis our study, and report classification experiments in Section~\ref{sec:experiments}. In Section~\ref{sec:demo}, we report the demo we developed using our models. Finally, we conclude and point to possible research directions for future work in Section \ref{sec:conclusion}.

\section{Related Work}
\label{sec:related_work}

Gender inference is a well-studied problem in English. \citet{Liu2013WhatsIA} present a dataset of 13K gender-labeled Twitter users and propose the use of first names as features for gender inference. Screen\_names, full names, user descriptions, and tweets have also been used as features for gender inference \cite{burger-etal-2011-discriminating}. \citet{rao2010Classifying} use stacked SVMs for identifying gender and other latent attributes of Twitter users. Semi-supervised methods that exploit social networks have also been used for gender classification \cite{li-etal-2016-semi}.

Gender inference has also received attention for a few other languages. \citet{sakaki-etal-2014-twitter} combine the output of text processor and image processor to infer the gender of Japanese Twitter users. \citet{taniguchi-etal-2015-weighted} propose a hybrid method that uses logistic regression to combine text and image features. \citet{ciot-etal-2013-gender} label 1000 users for gender in each of the following languages:  Japanese, Indonesian, Turkish, and French. The authors use Support Vector Machines (SVMs) for classification. \citet{sezerer-etal-2019-turkish} present a dataset consisting of 5.5K Twitter users labeled for their gender. 
\citet{9023806} proposes clustering-based approaches for demographic analysis to support advertising campaigns. Very recently \citet{liu2021comparative} provided a large-scale study that investigate different inference techniques (e.g., classic machine learning to deep learning models) using Twitter data. The authors highlight that a simpler model performs well to infer age, however, sophisticated models (e.g., sentence embeddings) are important for gender. 

For Arabic, on the other hand, work is relatively less explored. \citet{malmasi-2014-data} use first names to classify the gender of Arabic, German, Iranian and Japanese names. \citet{ELSAYED2020159} uses neural networks to differentiate male and female authors of tweets in Egyptian dialect. \citet{HUSSEIN2019109} use classical machine learning classifiers such as Logistic Regression and Random Forest classifiers to identify gender in Egyptian tweets. \citet{habash-etal-2019-automatic} use deep learning for gender identification and uses Machine Translation for reinflection. \citet{Bsir2018Enhancing} use the gated recurrent unit (GRU) for gender identification in Facebook and Twitter posts. \citet{zaghouani2018arap} collect a corpus of 2.4M multi-dialectal tweets from 1600 accounts that are tagged for gender, age, and language.



Our work differs from previous work on gender analysis and inference for Arabic in a number of ways {\em(i)} it uses a much bigger dataset for male and female users; {\em(ii)} it has no bias towards a specific country as it covers users from all Arab countries; {\em(iii)} it uses a generic method for collecting users and their names as opposed to starting with a specific list of names, which can be skewed towards some countries or cultures; {\em(iv)} in addition to gender inference, we perform a thorough analysis of gender differences in their profile descriptions, topics of interest, the profession gender gap among other things.

\section{Dataset}
\label{sec:dataset}

\subsection{Arabic Background}
\label{ssec:arabic_back}
In Arabic, typically nouns and adjectives have gender markers such as Taa Marbouta letter ``\<ة>'' as a feminine (f) suffix, and in case of absence, they can be considered as masculine (m). 
There are special cases where a word can have the feminine marker and it's gender is unknown (e.g., \<داعية> - religious scholar (m and f)). Also, there are some cases where words are feminine without explicit gender markers (e.g., \<أنثى، بنت> - female, girl). Except for some special cases, converting gender from masculine to feminine can be done by appending the Taa Marbouta suffix ``\<ة>'', e.g., words like \<مديرة، شاعرة> (manager(f), poet(f)) are the feminine forms of \<مدير، شاعر> (manager(m), poet(m)) in order.

It's widely observed that many users on Arabic Twitter describe themselves in the user description field in their profiles. This description expresses several identity features such as: nationality (NAT), profession or job (PROF), interest (INT), social role (SOC), religion (RELIG), ideology (IDEO) among others. We provide a few examples in Table~\ref{tab:user-description}.

\begin{table}[h]
    \resizebox{\linewidth}{!}{%
	\centering
	\begin{tabular}{rll}
	\toprule
		\textbf{Description} & \textbf{Translation} & \textbf{Class} \\ \midrule
		\<عراقي وأفتخر> & Iraqi (m) and proud & NAT\\
		\<مواطنة سعودية> & Saudi citizen (f) & NAT\\
		\rowcolor{LightCyan}
		\<طبيبة أسنان> & Dentist (f) & PROF \\
		\rowcolor{LightCyan}
		\<طالب دكتوراه> & PhD student (m) & PROF \\
		
		\<عاشقة الطبيعة> & Nature lover (f)& INT\\
		\<مهتم بأخبار التقنية> & Interested (m) in IT news& INT\\
		
		\rowcolor{LightCyan}
		\<زوجة وأم> &  Wife and mother & SOC\\
		\rowcolor{LightCyan}
		\<شاب متفائل> & Optimistic young man& SOC\\
		
		\<مسلم وأفتخر> & Muslim (m) and proud & RELIG\\
		\<مسيحية عربية> & Arab Christian (f) & RELIG\\
		
		\rowcolor{LightCyan}
		\<سياسي معارض> & Opposition politician (m) &  IDEO\\
		\rowcolor{LightCyan}
		\<ليبرالية أحب بلدي> & Liberal (f), love my country & IDEO\\
		\bottomrule
	\end{tabular}
	}
	\caption{Examples of user description with gender (m/f) and identity features (class).}
	\label{tab:user-description}
\end{table}

\subsection{Data Collection}
\label{ssec:data_collection}
For the data collection, we used Twitter API to crawl Arabic tweets using a language filter set to Arabic (``lang:ar"), back in January 2018. We collected data in two phases. \textit{First}, we collected 4.35M tweets (\textit{termed as former set}), which covers tweets from 2008 until the date of collection.\footnote{Note that our data collection might not consist of all of the tweets posted on Twitter during this period, which is because Twitter's free API has a limit.} Using this dataset we developed a word list using a gender marker (see Section \ref{ssec:creating_word_list}). In the \textit{second} phase, we collected additional 100M millions tweets (\textit{termed as later set}), dated from 2018 to 2020, to develop final annotated dataset (see Section \ref{ssec:gender_location_data_annotation}). The purpose of the \textit{former set} of tweets was to create a gender marker word list, the purpose of the \textit{later set} of tweets was to create a large annotated dataset with gender and location labels. We used such an approach to avoid any biases that may appear due to the word list selection. 

 


\section{Annotation}
\label{sec:annotation}

\subsection{Creating Word List with Gender Marker}
\label{ssec:creating_word_list}
For the annotation we first created a word list of gender markers. In order to do that we first extracted all profile information of users who posted these tweets. From the user description, we obtained a list of all first words that users used to describe themselves.\footnote{First word is a very strong signal in identity description and can be mapped to gender easily.} We obtained a unique list of 10K words. We then excluded words that appeared only once, which resulted in a  list of $\sim$2500 words out of 10K. We used the publicly available Farasa tool \cite{darwish2016farasa} to initially detect the gender of each word in the list. Then, a native speaker revised gender information and provided both the masculine and feminine word forms and their different writings to have better coverage. For example, for the feminine form ``\<محامية> - lawyer (f)", the masculine form and its different writings ``\<محامي، محامى، محامٍ> - lawyer (m)" were also added if they did not appear in the word list.
The final gender marker word list contains 713 words, in which 56\% of them indicate masculine and 44\% indicate feminine gender.\footnote{Words like \<شخص، كاهن، زول> (person, priest, man) have no corresponding feminine words.} The list can be found with our publicly released dataset.

\subsection{Gender and Location Annotation}
\label{ssec:gender_location_data_annotation}
For gender and location annotation, we first collected another set of 100M tweets, \textit{the later set}, which dated from 2018 to 2020.

\begin{figure}[t]
	\begin{center}
		\includegraphics[scale=0.65]{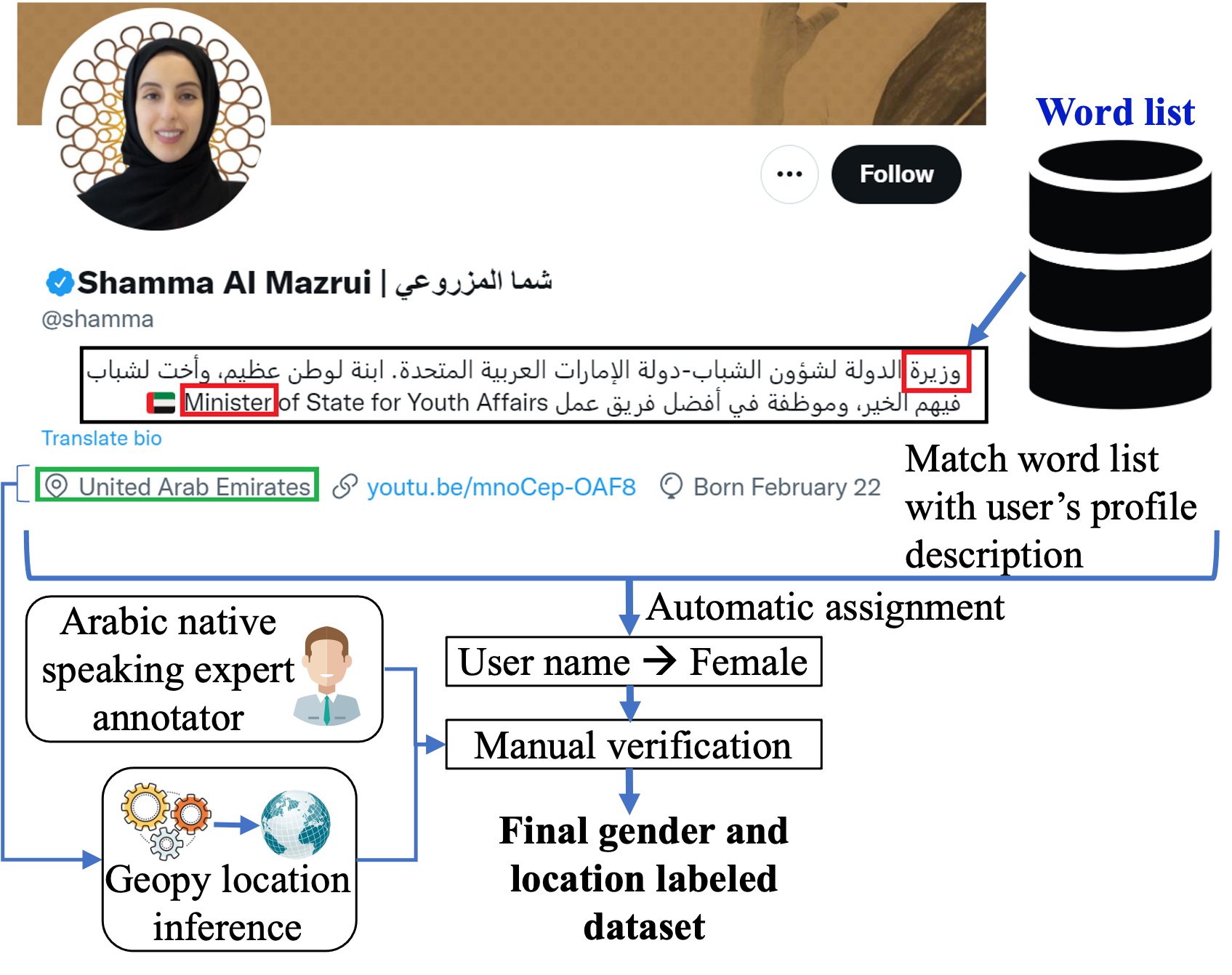} 
		\caption{Our pipeline to develop \textbf{ArabGend} -- labeling gender and location.
		}
		\label{fig:user_matching_wordlist}
	\end{center}
\end{figure}

\paragraph{Gender:}We annotated 100M tweets with gender and location information in several steps. We used the word list, discussed in the previous section, and matched the words at the beginning of each user's profile description. The matching approach resulted to assign a gender label to 167K users. We could not able to assign the gender label for the rest of the users due to the mismatch between our created word list, and the empty user's profile description. We then manually revised the assigned gender labels of these 167K users by a native Arabic-speaking expert annotator. In Figure \ref{fig:user_matching_wordlist}, we present \textit{ArabGend} development pipeline that demonstrates how user profile appears, how we use profile description with the word list to assign gender marker, and location information to assign specific location. Note that we developed the \textit{word list}, highlighted in blue, at the first phase of our dataset development, as discussed in Section \ref{ssec:creating_word_list}. In this profile, user location is clearly visible, however, this is not always the case for which location inference is needed.



\paragraph{Location:} Out of these 167K users we extracted 28K unique locations, which are then mapped into Arab countries with geographic location information using \textit{GeoPy toolkit}.\footnote{\url{https://pypi.org/project/geopy/}, It is a python client for several geocoding web services including Nominatim (\url{https://nominatim.org}), which uses OpenStreetMap data to find location.} Similar to gender annotation, the output of GeoPy is then manually revised by the same annotator. 
The annotation process resulted in to identify the countries for 92K users (55.08\% of all users) out of 167K users. We could not identify the rest as many of user locations as they were either empty (38\%) or cannot be mapped to a specific country (6.92\%).

\paragraph{Removing Ambiguous and Inappropriate Accounts}
The manual annotation process consists of another step to remove ambiguous, adult, and spam accounts. Typically Arabic words are written without diacritics which causes ambiguity in many cases, e.g., the word \<مدرسة> can be interpreted as Teacher (f) or School. As we are interested in collecting personal accounts using their profile description, therefore, we excluded organizations' accounts from our data collection. Also, there are some titles that can be used to describe males and females, which we removed. For example, \<دكتور، مدير> (Doctor, Manager) are used for both genders.

To filter adult and spam accounts we used the publicly available APIs from ASAD system
\cite{hassan2021asad}.\footnote{\url{https://asad.qcri.org}} Based on the classified output from ASAD and a manual inspection during the annotation process, we removed those accounts. We use the term \textit{appropriateness} to refer to the labels adult and spam in the rest of the paper.  

In this phase, after filtering non-personal and inappropriate accounts, we ended up with 166K users (80\% are males and 20\% are females) out of 167K users.

\subsection{Annotation quality}
In order to assess the quality of the annotation, we additionally manually annotated 500 users' accounts. We selected a random sample of 500 users and then manually assigned gender labels by checking their accounts on the Twitter platform. Agreement with manual annotation was $\sim$99\%. Similarly, for location, we randomly selected another sample of 500 unique user locations and checked their mappings to countries. The accuracy was 98\%, which indicates annotation quality is very high for gender and location labels. Note that, Twitter user locations are typically noisy and mapping them to countries is not always trivial.

\begin{table}[tbh!]
	\centering
	\scalebox{1.0}{
	\begin{tabular}{l|r|r}
	    \toprule
		\multicolumn{1}{c}{\textbf{Accounts}} & \multicolumn{1}{c}{\textbf{Count}} & \textbf{User Loc.} \\ \midrule
		Male & 133,192 (80.0\%) & 75,539 (81.5\%) \\
		Female & 33,348 (20.0\%) & 17,115 (18.5\%) \\
		\textbf{Total} & 166,540 (100\%) & 92,654 (56.0\%)\\
		\bottomrule
	\end{tabular}
	}
	\caption{Statistics of the dataset.}
	\label{tab:corpus-stats}
\end{table}

\begin{table}[tbh!]
	\centering
    \resizebox{\linewidth}{!}{%
	\begin{tabular}{l|l|l|ll}
	\toprule
		\textbf{User Name} & \textbf{Description} & \textbf{User Loc.} & \textbf{G} & \textbf{C}\\ \midrule
		\<صفية الشحي> & \<إعلامية - كاتبة> & UAE - Dubai & F & AE\\
		(Safia Alshehi) & (journalist (f) and writer (f)) & & & \\
		
		\rowcolor{LightCyan}
		Ahmed Azhar & \<إنسان بسيط جدا > & \<جدة> & M & SA \\
		\rowcolor{LightCyan}
		& (very simple person (m)) & (Jeddah)& & \\
		\bottomrule
	\end{tabular}
	}
	\caption{Annotation examples: Description was mapped to Gender (G), and User Loc. was mapped to Country (C)}
	\label{tab:examples}
\end{table}

\subsection{Statistics}
In Table~\ref{tab:corpus-stats}, we report number of final male and female accounts and percentage of successful mappings of user locations to countries for both genders. 
According to a report from the World Bank in 2015,\footnote{\url{https://blogs.worldbank.org/ar/arabvoices/ten-facts-about-women-arab-world}} the gender gap in Middle East and North Africa region can reach to 34\% in internet usage. This gap comes second after the largest gender gap in Sub-Saharan Africa region (45\%). Further, while 52\% of females (91M) have mobile phones, this ratio increases to 56\% for males with additional 8M male users. These factors can explain the less presence of female users on Twitter as shown in our study. In Table~\ref{tab:examples}, we present some annotation examples from our dataset. We use ISO 3166-1 alpha-2 for country codes.\footnote{ \url{https://en.wikipedia.org/wiki/List_of_ISO_3166_country_codes}}

\begin{figure}[!tbh]
	\begin{center}
		\includegraphics[scale=0.45]{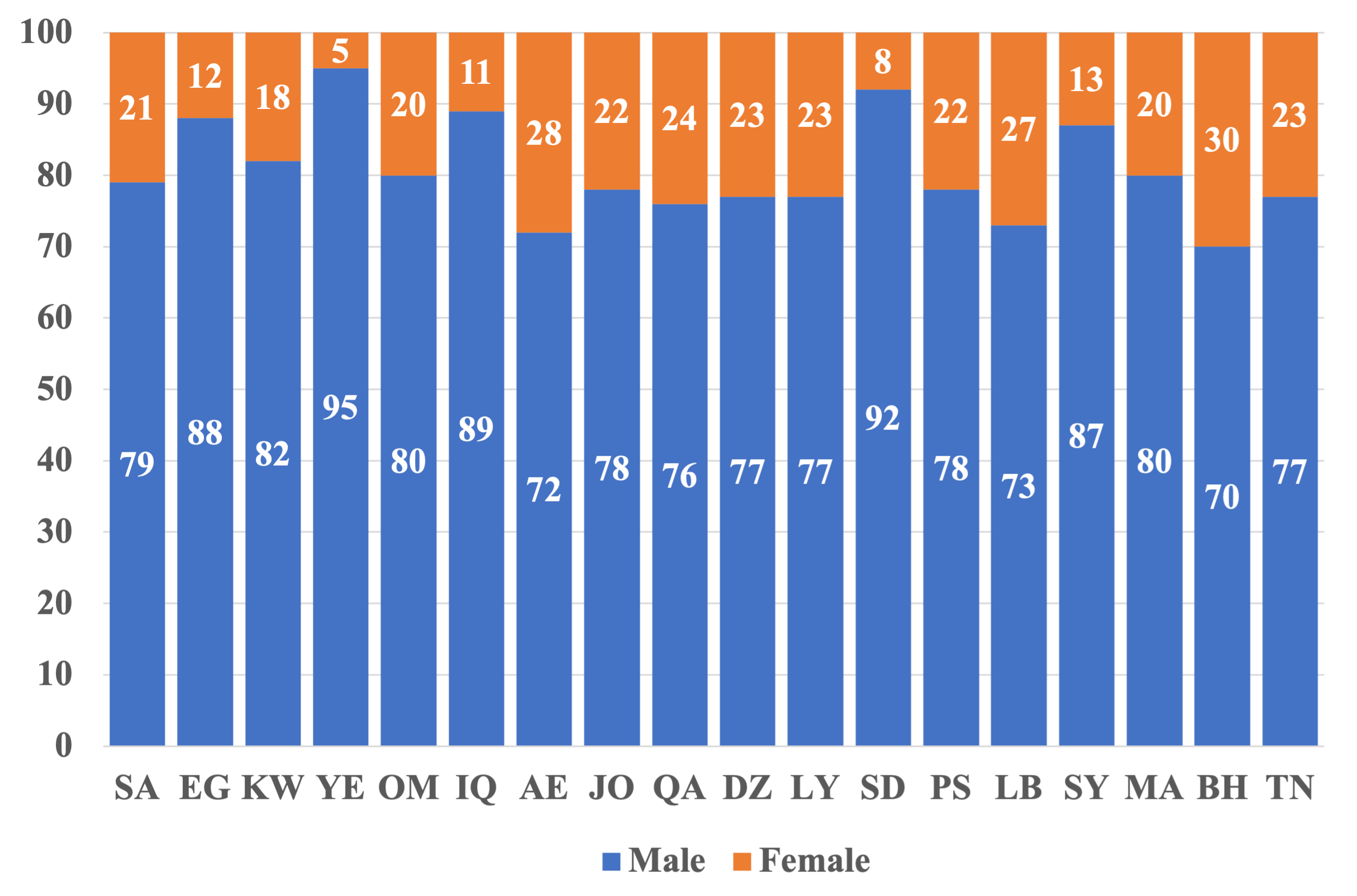} 
		\caption{Gender distribution in Arab countries}
		\label{fig:country-gender}
	\end{center}
\end{figure}

\begin{figure}[!tbh]
	\begin{center}
		\includegraphics[scale=0.50]{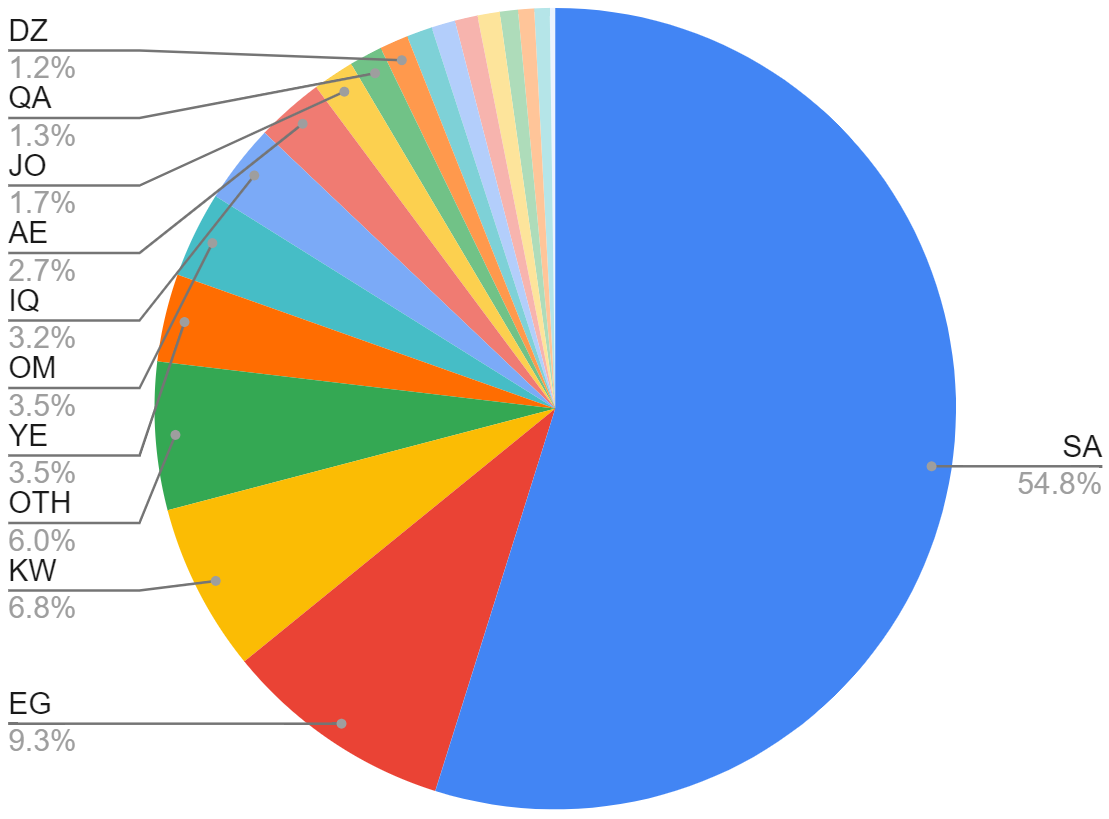} 
		\caption{Country distribution of Twitter accounts.}
		\label{fig:user-country}
	\end{center}
\end{figure}

\section{Analysis}
\label{sec:analysis}

\subsection{Gender and Location Distribution}
In Figure \ref{fig:country-gender}, we present gender distribution of Twitter users in Arab countries. We observe that the top three countries that have higher percentages of female users for BH (Bahrain), AE (United Arab Emirates) and LB (Lebanon) are 30\%, 28\% and 27\%, respectively. 
The lowest percentages of female users from YE (Yemen), SD (Sudan) and IQ (Iraq) are 5\%, 8\% and 11\%, respectively. 

In Figure \ref{fig:user-country}, we present country distribution of all accounts in our dataset. We observe that more than half of Twitter users are from SA (Saudi Arabia) and 70\% of accounts are from Gulf region (countries: SA, KW, OM, AE, QA and BH) followed by accounts from EG, YE, etc. We mapped user locations to OTHER (OTH) for the countries that are outside Arab World. 
They represent 6\% of all user locations. Top five countries that are outside Arab World include US, GB, TR, DE and FR in order.

We found that the dataset has 1,495 verified accounts, out of which 90\% are male and 10\% are female. Such a number represents 1\% and 0.45\% verified male and female accounts, respectively.


\subsection{User Engagement}
\begin{figure*}[h]
	\begin{center}
		\includegraphics[scale=0.75]{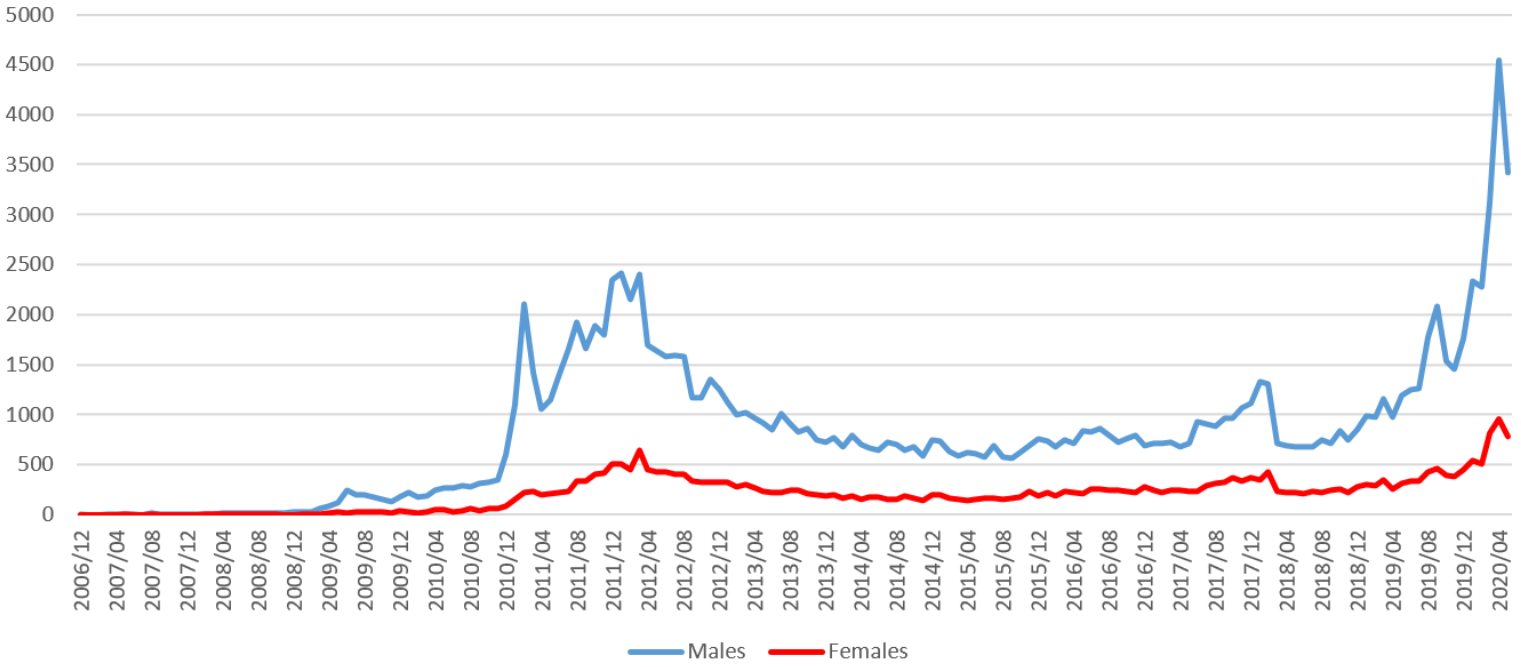} 
		\caption{Distribution of Twitter joining date}
		\label{fig:user-created-at}
	\end{center}
\end{figure*}

\begin{figure*}[h]
	\begin{center}
		\includegraphics[scale=0.75]{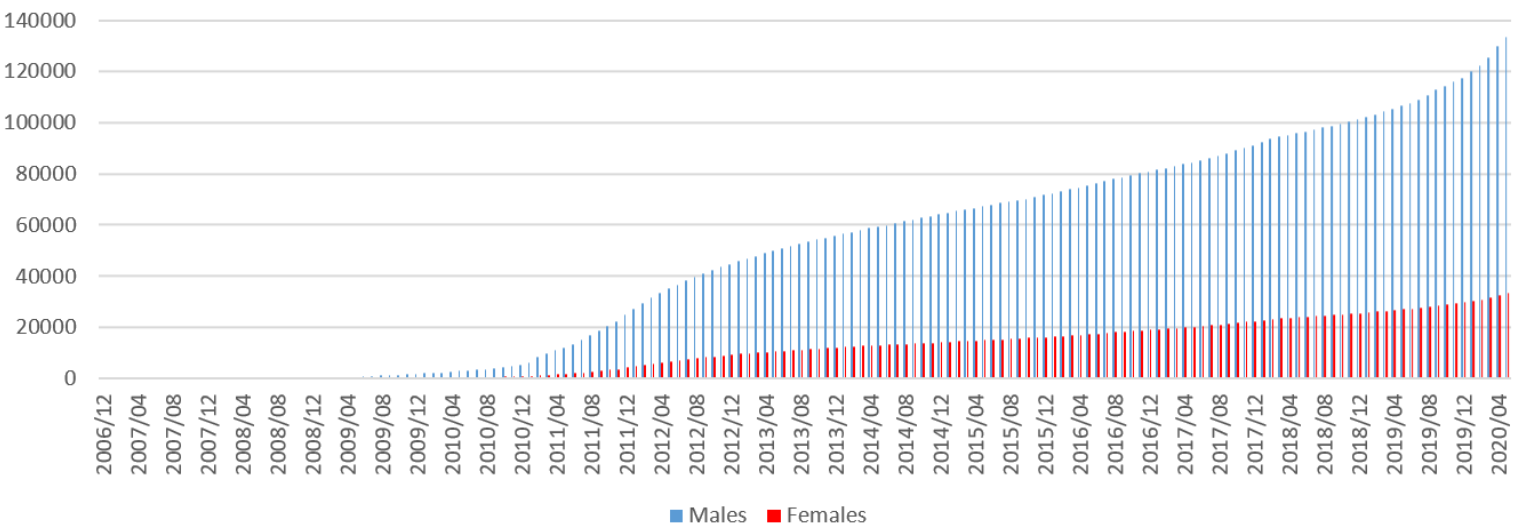} 
		\caption{Accounts distribution over time}
		\label{fig:user-created-at-cumualtive}
	\end{center}
\end{figure*}

We extracted the date of joining Twitter for all accounts to study their engagement with Twitter. As shown in Figure \ref{fig:user-created-at}, we can see that many accounts joined Twitter between 2010 and 2012, then the number of users who joined Twitter between 2013 and 2018 was almost stable for male and female accounts. Starting from 2019, there was an increasing number of joining users. We notice that there is a slightly increasing number of female accounts who join Twitter over time, however, Twitter was always dominated by male accounts and the gap between the two genders seems to increase in the future as shown in the cumulative chart in Figure \ref{fig:user-created-at-cumualtive}.

\subsection{User Connections}
\begin{figure}[h]
	\begin{center}
		\includegraphics[scale=0.75]{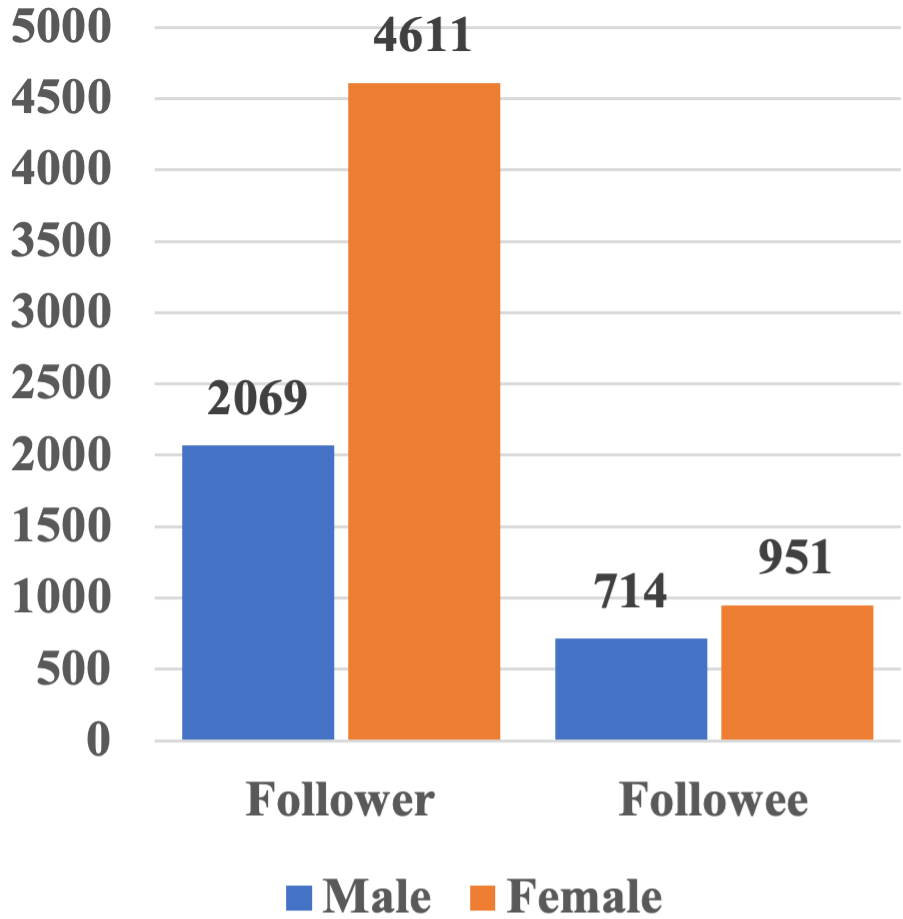} 
		\caption{Followers and followees distribution}
		\label{fig:user-followers}
	\end{center}
\end{figure}

Figure \ref{fig:user-followers} shows an average number of followers and followees (friends) of male and female accounts in our dataset. We can see that on average, female accounts tend to attract more followers than males (more than double). Further, females have $\sim$30\% more friends than males which may indicate that females prefer to have a larger community and friends than males on Twitter.

\subsection{Person Names}
A person's name is a very important feature in identifying gender. To understand the demographics of Twitter users, prior studies have been using a seed list of names to collect male and female accounts. \citet{mislove2011understanding} used the most common 1000 male and female names in the US to collect Twitter user information. Such an approach, i.e., using a pre-specified list of person names,  can create bias in the resulting data collection. In our study, we attempted to follow a different approach to avoid such a bias. We created \textit{initial} dataset to create word list, and used a different set (i.e., the \textit{later} 100M) to create the final list. 
We further normalized the names by removing diacritics, mapping Alif shapes, Taa Marbouta and Alif Maqsoura letters to plain Alif, Haa, and Yaa letters respectively, and mapping decorated letters to normal letters.


From the obtained lists, we can extract names that can be used for both genders when they are written in Arabic (e.g., \<نور، صباح، شمس> - Nour, Sabah, Shams), or due to transliteration ambiguity, e.g., the names \<علاء>(m) and \<آلاء>(f) both are transliterated to ``Alaa'', also \<أمجد>(m) and \<أمجاد>(f) have the same transliteration ``Amjad''. 

\begin{figure}[h]
	\begin{center}
		\includegraphics[scale=0.4]{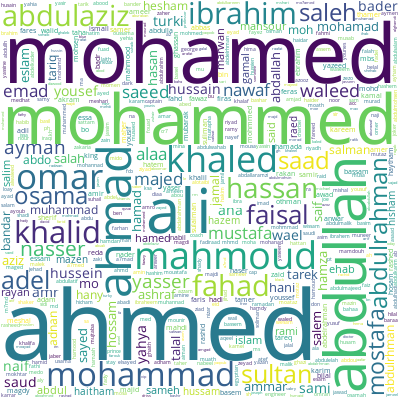} 
		\caption{Common Arabic male names in English}
		\label{fig:names-m-e}
	\end{center}
\end{figure}

\begin{figure}[h]
	\begin{center}
		\includegraphics[scale=0.4]{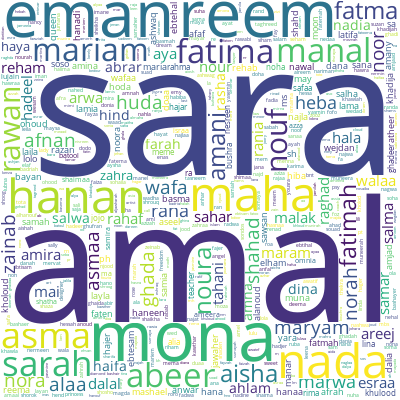} 
		\caption{Common Arabic female names in English}
		\label{fig:names-f-e}
	\end{center}
\end{figure}

In Figures \ref{fig:names-m-e} and \ref{fig:names-f-e}, we show the most common male and female names written in English. Mostly, they have similar distribution as their Arabic counterparts with different ways of transliteration.

The full list of male and female names written in Arabic and English will be available in our dataset. The extracted list of person names can be used for further analysis.




\subsection{Interests According to User Description}
\begin{figure}[h]
	\begin{center}
		\includegraphics[scale=0.4]{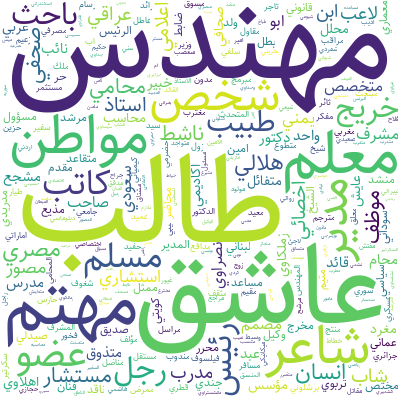} 
		\caption{Description of male accounts. The top five include engineer, student, lover, interested (in), and teacher.}
		\label{fig:names-desc-m}
	\end{center}
\end{figure}

\begin{figure}[h]
	\begin{center}
		\includegraphics[scale=0.4]{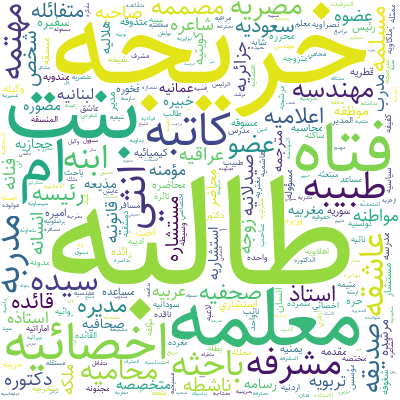} 
		\caption{Description of female accounts. The top five includes graduate, student, daughter, teacher, and girl.}
		\label{fig:names-desc-f}
	\end{center}
\end{figure}

In Figures \ref{fig:names-desc-m} and \ref{fig:names-desc-f}, we present most common words used in user description for males and females in order. This gives an indication about jobs and interests for both genders. We can see that females tend to describe their social role (e.g., \<بنت، أم، فتاة، صديقة> - daughter, mother, girl, friend) more than males. For comparison, while more than 1000 female accounts describe themselves first as \<أم> (mother), less than 200 accounts describe themselves as \<أب> (father). We can also see that a good portion of Twitter users is young (e.g., \<طالب، فتاة، خريجة> - student, young woman, graduate) as opposed to few accounts who describe themselves as \<متقاعد> (retired). From our analysis, we observed that self-description can be used to predict the age group of Twitter users. We leave this for future work.

\subsection{Topics of Interest}
\begin{figure}[h]
	\begin{center}
		\includegraphics[scale=0.5]{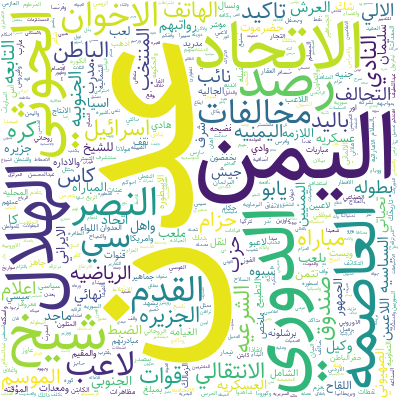} 
		\caption{Most common words in males tweets.}
		\label{fig:text-m}
	\end{center}
\end{figure}

\begin{figure}[h]
	\begin{center}
		\includegraphics[scale=0.5]{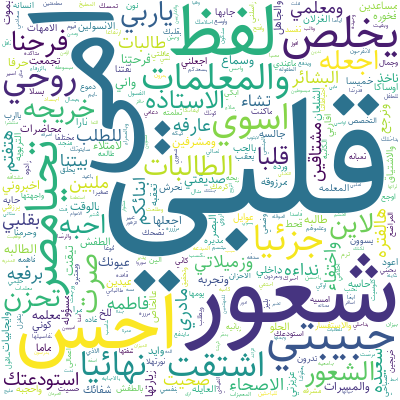} 
		\caption{Most common words in females tweets.}
		\label{fig:text-f}
	\end{center}
\end{figure}

In Figures \ref{fig:text-m} and \ref{fig:text-f}, we present the common distinguishing words in tweets written by male and female accounts in our dataset. We used the valence score formula as shown in Equation \ref{eq:valence}, discussed in \cite{conover2011political,chowdhury2020multi} with 0.5 as a threshold to obtain these words. 

\begin{equation}
\label{eq:valence}
    \vartheta (x, L_i) = 2 * \frac{\frac{C(x|L_i)}{T_{L_i}}}{\sum_{l}^{L} C(x|L_l) } -1
\end{equation}
\noindent where $C(.)$ is the frequency of the token $x$ for a given category $L_i$. $T_{L_i}$ is the total number of tokens present in that category. In $\vartheta(x) \in [-1, +1]$, the value $+1$ indicates the use of the token is significantly higher in the target category than the other categories. Here the categories are male and female.

While tweets from males have many words related to politics (e.g., \<اليمن، الإخوان> - Yemen, Muslim Brotherhood) and sports (e.g., \<الدوري، الهلال> - league, Hilal club), tweets from females have many words related to family and society (e.g., \<أمي، أبناء، معلمات، زميلات> - my mother, children, teachers, colleagues) and feelings\newline
(e.g., \<قلبي، شعور، حبيبتي> - my heart, feeling, my love).

\subsection{Gender Gap in Professions}
We can observe from Figures \ref{fig:names-desc-m} and \ref{fig:names-desc-f} that the most frequent profession for males was \<مهندس> (engineer) while it was \<معلمة> (teacher) for females. In Table \ref{tab:jobs}, we report the distribution of some professions for male and female accounts in different domains. We observed that the Sports domain is overwhelmingly dominated by males, and other domains (e.g., Management, Software, Health, etc.) have less representation of females (percentages are from 9\% to 20\%). The best domain that has a good representation of females is the Translation domain with a percentage of 36\%.

According to the World Bank's report in June 2020,\footnote{\url{https://data.worldbank.org/indicator/SL.TLF.CACT.FE.ZS?locations=ZQ}} the labor force participation rate of females in the Middle East and North Africa region is around 20\% with a slight improvement from 17.4\% in 1990. Our study supports this report by showing that females are less represented in many job domains, and participation rates can be roughly quantified in different sectors of job markets. 
The same report also mentions that only 11\% of females hold managerial positions compared to the world average of 27\%.\footnote{\url{www.dw.com/ar/}, \url{shorturl.at/vLOQT}} The ratio of female managers to all managers in our dataset is 9\% based on profile self-disclosure. 

\begin{table}[h]
	\centering
	\scalebox{1.0}{
	\begin{tabular}{rlcrrl}
	\toprule
		\textbf{Prof.} & \textbf{Translation} &  \textbf{G} & \textbf{Freq.} & \textbf{\%} & \textbf{Domain}\\ \midrule
		\<لاعب> & player  &m & 1,096 & 98 & Sport\\
		\<لاعبة> &  & f & 19 & 2 & \\
		
		\rowcolor{LightCyan}
		\<مهندس> & engineer & m & 6,619 & 94 & Engineering\\
		\rowcolor{LightCyan}
		\<مهندسة> &  & f & 404 & 6 & \\
		
		\<مدير> & manager & m & 2,982 &91 & Management\\
		\<مديرة> & & f& 286 &9 & \\
		
		\rowcolor{LightCyan}
		\<مبرمج> & programmer & m& 153 &91 & Software\\
		\rowcolor{LightCyan}
		\<مبرمجة> &  & f& 16 &9 & \\
		
		\<محاسب> & accountant & m& 580 & 90& Finance\\
		\<محاسبة> & &  f& 61 &10 & \\
		
		\rowcolor{LightCyan}
		\<طبيب> & doctor & m & 2,265 & 80 & Health \\
		\rowcolor{LightCyan}
		\<طبيبة> &   & f &577 & 20 & \\
		
		\<مترجم> & translator & m & 177 &64 & Translation\\
		\<مترجمة> & & f& 98 &36 & \\ \bottomrule
	\end{tabular}
	}
	\caption{Profession gaps examples}
	\label{tab:jobs}
\end{table}

\section{Experiments}
\label{sec:experiments}
For the classification experiments, we focused only on the gender inference and leave the location inference study as for a future study. We measure the performance of the classification models using accuracy (Acc), macro-averaged precision (P), recall (R) and F1 score. We use macro-averaged F1 score as primary metric for comparison in our discussion.

\subsection{Datasets} 
We used two datasets for training to provide a comparative study. We used our developed \textit{ArabGend} dataset only for training. We also used \textit{ARAP} dataset \cite{charfi2019fine}, which consists of 1600 Twitter accounts labeled for their gender along with country and language. We used half of the \textit{ARAP} dataset for training, and half for the evaluation. Hence, in our experiments, models are evaluated using half \textit{ARAP} dataset, which we considered as our test set. 


\subsection{Classification Models and Features} We used Support Vector Machines (SVMs) as our classifier. As features, we used character n-gram vectors weighted by term-frequency-inverse document term frequency (tf-idf). We experimented with different n-gram ranges. Only character [2-5] n-gram results are reported in this paper since they yielded the best results.

In addition to that we also varied what input the classifiers should have. We experimented with {\em(i)} a single tweet from each user, {\em(ii)} aggregate all tweets from a user, {\em(iii)} usernames of the Twitter users. We also experimented with by balancing the \textit{ArabGend} training set, to have same number males and females, to understand the affect on the performance of the classifiers. Since ARAP Tweet is balanced in terms of gender, hence, we do not apply any sampling to balance data any further. Since there was not significant improvement in performance after balancing with equal distribution, therefore, we do not report that results.


\subsection{Results}
In Table \ref{tab:experiments}, we report the classification results on ARAP test set. From the results, we observed that for both ARAP-Tweet data and our data, best results are obtained when usernames are used as opposed to aggregation of tweets or user descriptions. In general, aggregating tweets do not improve results in general by a significant margin. The usernames on our data have a significant performance improvement over all other settings, resulting in an F1 score of 82.1.


\begin{table}[h]
	\centering
	\scalebox{1.0}{
	\begin{tabular}{@{}llcccc@{}}
		\toprule
		\multicolumn{1}{c}{\textbf{Train Data}} & \multicolumn{1}{c}{\textbf{Features}} & \textbf{Acc.} & \textbf{P} & \textbf{R} & \textbf{F1} \\ \midrule
		Majority Baseline & & 53.3	& 26.7 &	50.0 &	34.8\\
		& Usernames & 67.2 & 67.1 &	66.8 & 66.8\\
		ARAP (Baseline) &  Description & 58.2	& 58.5 & 58.5 & 58.2\\
		& Tweets & 69.8 & 70.9 & 70.4 &	69.7\\
		& All Features & 59.9 & 65.3 & 61.6 & 57.9\\ \midrule
		& Usernames & \bf{82.4}	& \bf{82.7} & \bf{82.0} &	\bf{82.1}\\
		ArabGend    &    Description & 64.1	& 65.4	& 62.7 & 61.8\\
		& Tweets & 63.1 & 62.9 & 62.9 &	62.9\\
		& All Features & 78.0	& 80.2 & 77.1 &	77.1\\ \bottomrule
	\end{tabular}
	}
	\caption{Performance on ARAP test data}
	\label{tab:experiments}
\end{table}

\subsection{Additional Experiments}
\label{sec:aAdditional_exp}
\paragraph{\textbf{Predicting Gender from Profile Images}}
To evaluate the efficiency of using tools that detect gender from profile images, we user Gender-and-Age-Detection tool\footnote{\url{https://github.com/smahesh29/Gender-and-Age-Detection}} on ARAP test set. It uses deep learning to identify the gender and age of a person from face image, in which model was trained on $\sim$27K images from Flickr (Adience dataset) \cite{levi2015age}. Accuracy of this tool was 64\%.\footnote{Some images are hard for gender prediction, e.g., flag, natural scene, incomplete face, kid image, cartoon, mixed, etc.}

\begin{figure}[h]
	\begin{center}
		\includegraphics[scale=0.8]{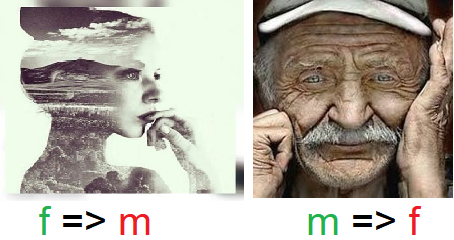} 
		\caption{Profile image classification errors of Gender-and-Age-Detection tool.  
		}
		\label{fig:profile-image-errors}
	\end{center}
\end{figure}

For comparison, we manually annotated the same ARAP test set for gender prediction using profile images and the accuracy was 81\%. This shows that profile image can be one of the powerful features to predict gender. It is worth to mention that 87\% of the package errors are due to misclassification of female users as males. Some examples of error\footnote{Non-personal images are shown for privacy protection.} are shown in Figure \ref{fig:profile-image-errors}. We leave integrating profile images with textual features for future work. 

\paragraph{\textbf{Predicting Gender from Friends Network}}
Homophily (meaning love of the same) is a tendency in social groups for similar people to be connected together \cite{mcpherson2001birds}. Homophily has predictive power in social media \cite{bischoff2012we}. We anticipated that female users on Twitter tend to have more female friends than male users and vice versa. 

To experiment this assumption, we collected a list of up to 100 friends\footnote{We used twarc API to get list of friends.} for all accounts in the ARAP test set, and from their usernames, we used our classifier to predict their gender. We experimented with different thresholds on ratio of predicted male to predicted female friends to decide gender of our target users. The best results were obtained when 1/3 of friends of an account are predicted as females. In these cases, we propagated the label ``female'' to the account and propagated ``male" otherwise. By doing so, we could achieve 56\% accuracy. This shows that gender distribution of friends network has limited impact on determining gender of a user.

We also explored if information about friend's gender can improve the performance of the model from the earlier section. We adopt the following procedure: if the classifier is not confident that the instance is male, we apply the threshold technique above and take the classifier's predicted label otherwise. By doing this, we were able to improve the performance from 82.1\% to 82.9\% indicating that friend's gender might be helpful in cases where the classifier is not confident.
However, obtaining a list of friends for all accounts in our dataset needs a significant amount of time. This limits the usage of friends' gender in cases where fast response is needed.

\begin{figure*}[!h]
	\begin{center}
		\includegraphics[scale=0.60]{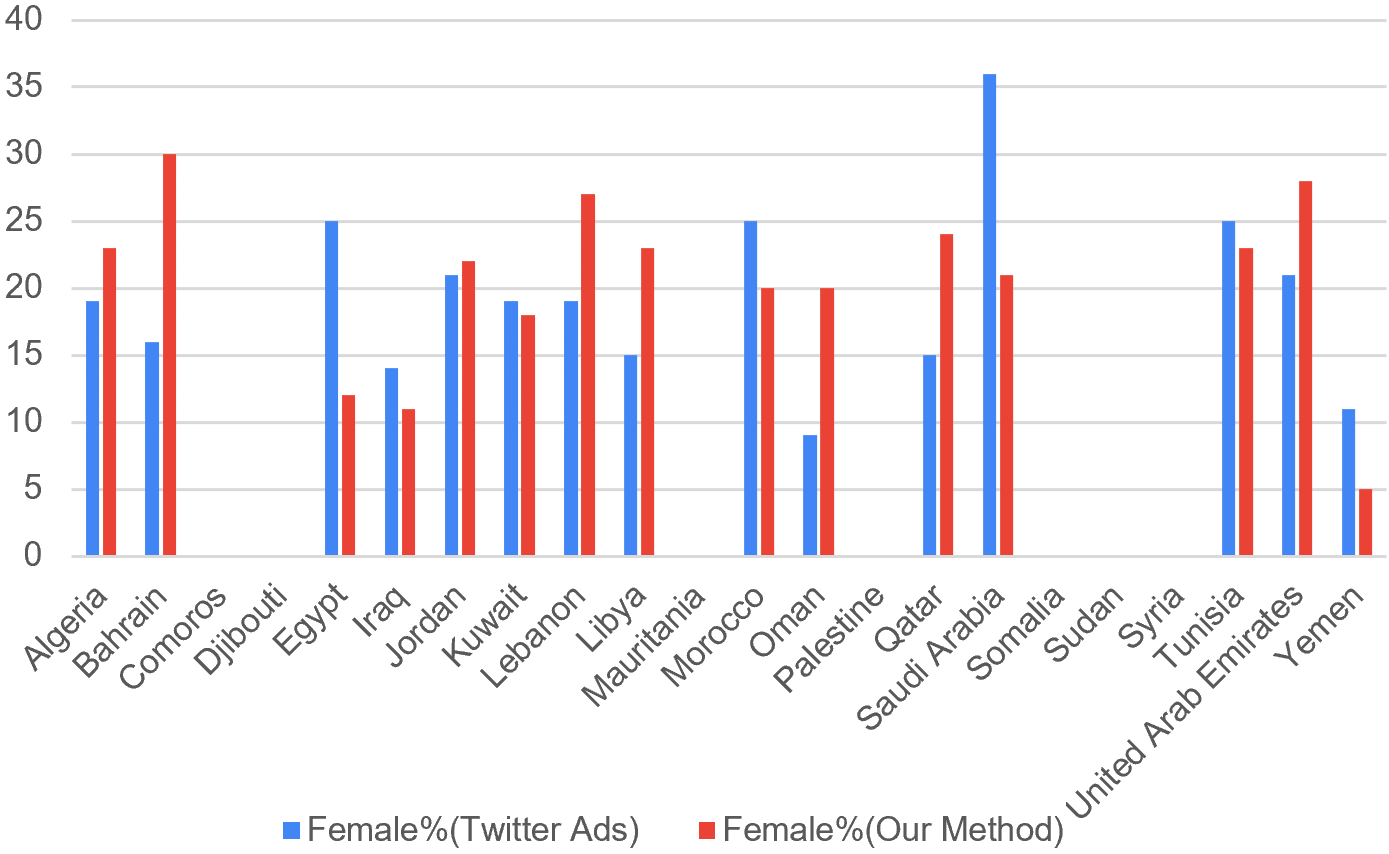} 
		\caption{Distribution of female accounts}
		\label{fig:female-twitter-ads}
	\end{center}
\end{figure*}
\begin{figure*}[!tbh]
	\begin{center}
		\includegraphics[scale=0.3]{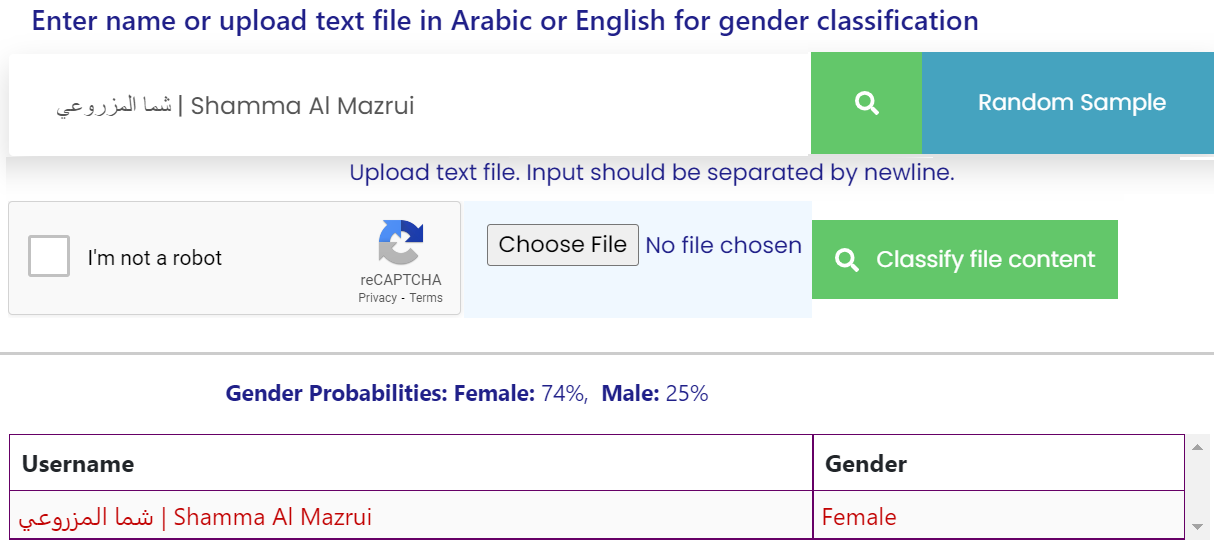} 
		\caption{Demo interface for gender inference using our proposed models. 
		}
		\label{fig:gender_inf_demo}
	\end{center}
\end{figure*}

\paragraph{\textbf{Comparison with Twitter Ads API}}
\label{sec:comparison}
Advertisers on Twitter can target their campaigns based on geo-location, gender, language, and age. Twitter uses the gender provided by people in their profiles, and extends it to other people based on account likeness.\footnote{ \url{https://business.twitter.com/en/help/campaign-setup/campaign-targeting/geo-gender-and-language-targeting.html}} We used Twitter Ads API to get total number of users in all Arab countries and their gender distribution.\footnote{Twitter Ads info are unavailable for some countries.}

Figure \ref{fig:female-twitter-ads} shows  distribution of female users as obtained from Twitter Ads and our method. Although there are some differences between the two methods, the average percentages of female users are similar (19\% using Twitter Ads vs. 20\% using our method). This can show that our method is close to Twitter Ads for gender prediction of users although Twitter has much larger information to use. We should take into account that Twitter Ads results may have limitations in terms of accuracy. 


\section{Demo} 
\label{sec:demo}
Using the developed model, we also built a demo that takes a person's name written in Arabic or English and predicts a gender label with probabilities. 
The demo can be accessed using the link: \url{http://hidden-for-blind-review.com}. A screenshot of the demo is presented in Figure \ref{fig:gender_inf_demo}.

\section{Conclusion}
\label{sec:conclusion}
In this paper, we have presented \textit{ArabGend}, a new dataset of Twitter users labeled for their gender and location. To the best of our knowledge, this is the largest Arabic dataset for gender analysis. We analyzed the characteristics of the users from a gender perspective. We identified key differences between male and female accounts on Arabic Twitter such as user connections, topics of interest, etc. We also studied the gender gap in professions and argued that results obtained from our dataset are aligned with recent reports from the World Bank and Twitter Ads information. We also showed that our dataset yields the best inference results on a publicly available test set. In the future, we plan to enhance our data collection method by considering gender markers in the whole user description and other profile fields.

\section*{Ethical Concern and Social Impact}
\paragraph{\textbf{User Privacy}}
For privacy protection and compliance with Twitter rules, we make sure that Twitter account handles and tweets are fully anonymized. We assign artificial user IDs for Twitter accounts 
and we share tweets by their IDs. We share lists of names written in Arabic and English as first names only. 
\paragraph{\textbf{Biases and Limitations}}
Any biases found in our dataset are unintentional, and we do not intend to cause harm to any group or individual. In our study, we tried to remove biases in data collection by providing all forms of male and female description words. But, because Twitter is widely used in some regions (e.g. Gulf) and less used in other regions (e.g. Maghreb), we acknowledge that our statistics and results may be less accurate for some Arab countries in the real world. However, they give rough estimates about the actual presence of users from those countries on Twitter. The bias in our data, for example towards a particular gender, is unintentional and is a true representation of users on Twitter as obtained also from Twitter Ads. Gender label (male/female) is extracted from the data and might not be a true representative of the users' choice.

Further, we heavily depend on users' self-disclosure which covers a portion of Twitter users but not all of them. In the future, we plan to consider better methods for data collection with greater diversity and coverage.

\bibliographystyle{ACM-Reference-Format}
\bibliography{bib/main}

\end{document}